\documentclass[12pt,a4paper,final]{iopart}

\usepackage{iopams}
\usepackage{graphicx}
\usepackage[breaklinks=true,colorlinks=true,linkcolor=blue,urlcolor=blue,citecolor=blue]{hyperref}
\usepackage{multirow}
\usepackage{cite}

\usepackage{color}
\definecolor{redcolor}{rgb}{0,0.,0.}
\def\Red#1{{\color{redcolor} #1}}

\newcommand{\matr}[1]{\mathbf{#1}} 

\renewcommand\appendix{\par
  \setcounter{section}{0}
  \setcounter{subsection}{0}
  \setcounter{figure}{0}
  \setcounter{table}{0}
  \renewcommand\thesection{Appendix \Alph{section}}
  \renewcommand\thefigure{\Alph{section}\arabic{figure}}
  \renewcommand\thetable{\Alph{section}\arabic{table}}
}

\begin{document}

\title[]{A complex network approach to stylometry}

\author{Diego R. Amancio}
\address{Institute of Mathematics and Computer Science\\
University of S\~ao Paulo, S\~ao Carlos, S\~ao Carlos, Brazil}
\ead{diego@icmc.usp.br, diego.raphael@gmail.com}


\begin{abstract}
Statistical methods have been widely employed to study the fundamental properties of language. In recent years, methods from complex and dynamical systems proved useful to create several language models. Despite the large amount of studies devoted to represent texts with physical models, only a limited number of studies have shown how the properties of the underlying physical systems can be employed to improve the performance of natural language processing tasks.  In this paper, I address this problem by devising complex networks methods that are able to improve the performance of current statistical methods. Using a fuzzy classification strategy, I show that the topological properties extracted from texts complement the traditional textual description. In several cases, the performance obtained with hybrid approaches outperformed the results obtained when only traditional or networked methods were used. Because the proposed model is generic, the framework devised here could be straightforwardly used \Red{to study similar textual applications
where the topology plays a pivotal role in the description of the interacting agents.}
\end{abstract}


\vspace{2pc}

\section{Introduction}

The human language represents a major factor responsible for the success of our species; and its written form is one of the main expression used to convey and share information. Owing to the ubiquity of language in several contexts, many linguistic aspects have been studied via the application of methods and tools borrowed from diverse scientific fields. As a consequence, several findings related to the origins, organization and structure of the language have been unveiled. One of the most fundamental patterns arising from statistical analysis of huge amounts of text is the Zipf's law, which states that the frequency of the words decreases inversely to their rank~\cite{zipf0,zipf,zipf1}. Other fundamental recurrent pattern is the Heap's law, which states that the vocabulary size grows slowly with the number of tokens of the document~\cite{heap1,heap2,heap3}. More recently, concepts from Physics have been applied to model several language features. For example, with regard to long-scale properties, concepts from dynamical systems have successfully been employed to compare the burstiness of the spatial disbribution of words in documents~\cite{katz,authorship}. In a similar fashion, the spatial distribution of words in documents and analogous systems have also been studied in terms of level statistics~\cite{level}, entropy~\cite{statistical} and intermittency measurements~\cite{imeas}

A well-known approach to study written texts is the word adjacency network model~\cite{ref1,wan,masucci}, which considers short-scale textual properties to form the networks. In this representation, relevant words conveying meaning are modeled as nodes and adjacency relationships are used to establish links. Using this model, several characteristics of texts and languages have been inferred from statistical analyses performed in the structure of networks~\cite{hliu}. Language networks have been increasingly employed to understand theoretical linguistic aspects, such as the origins of fundamental properties~\cite{origins} and the underlying mechanisms behind language acquisition in early years~\cite{acquisition}. In practical terms, networks have been applied in the context of machine translations~\cite{cnammt}, autommatic summarization~\cite{extractive}, sense disambiguation~\cite{navigli,unveiling}, complexity/quality analysis~\cite{complexity,forte} and document classification~\cite{mihalcea}. Despite the relative success of applying networks concepts to better understand language phenomena, in many real-world applications attributes extracted from networked models have not contributed to the advancement of the state of art.
For example, when one considers the document classification task, a strong dependency of network features on textual characteristics have been observed. However, the best performance are still achieved with traditional statistical natural language processing features.
In this context, the present study address this problem by devising methods that effectively take advantage of network properties to boost the performance of the textual classification task. Here I focus on the text classification based on stylistic features, where a text is classified according to stylistic marks left by specific authors~\cite{surveyaut} or literary genres~\cite{liter}. Upon introducing a hybrid classifier relying on the fuzzy definition of supervised pattern recognition methods~\cite{fuzzyknn}, I show that the performance of style-based classifications can be significantly improved when topological information is included in traditional models. \Red{Given the generality of the proposed method, the framework devised here could be applied to improve the characterization and description of many related textual applications.}



\section{Representing texts as networks} \label{represent}

There are several ways to map texts into networks~\cite{ref1,hliu,baroncheli,lantiq,extractive,patterns}. The most suitable form depends on the context of the application. In occasions where the semantics is relevant, the words sharing some semantical relationship are linked~\cite{unveiling,arenas,adv,highorder,complexity}. In a similar fashion, other semantic-based models link the words appearing in a given context~\cite{veronis,windows} (e.g. in the same sentence or paragraph). \Red{A general model for establishing significant ``semantical'' links between co-occurring elements was devised in~\cite{disen}. More specifically, the model considers the existence of a set of elements $V = \{v_1,\ldots,v_n\}$, which may occur in one or more sets of a given collection $\xi = \{S_1,\ldots,S_N\}$.
Given two elements $\alpha \in V$ and $\beta \in V$, the model computes the probability $p$ that more than $r$ sets in $\xi$ contain both elements $\alpha$ and $\beta$ as
\begin{equation}
    p_t = \sum_{j \geq r} p(j),
\end{equation}
where $p(j)$ is the probability that $\alpha$ and $\beta$ co-occur exactly $j$ times in the same set. Given a confidence level $p_0$, the strength of the link between $\alpha$ and $\beta$ is $s = \log(p_0/p_t)$.
}

%


\Red{In applications where the style (or structure) plays an important role, the links among words are established according to \emph{syntactical relationships}~\cite{extractive,patterns}}.
A well-known approach for grasping stylistic features of texts is the word adjacency model~\cite{wan,identification,machine}, which  basically connect adjacent words in the text. \Red{Differently from the model devised in~\cite{disen}, the word adjacency model captures the stylistic features of texts~\cite{voynich}}.
It has been shown that the adjacency model is able to capture most of the syntactical links with the benefit of being language independent. Despite being a simplification of the syntactical analysis, the adjacency model has been employed in several contexts because the topological properties of word adjacency and syntactical networks are similar. Such high degree of similarity can be explained by the fact that most of the syntactical links occurs among neighbouring words~\cite{patterns}. In the current paper, the traditional word adjacency representation was adopted.

\begin{figure}
  \centering
  \includegraphics[width=0.55\textwidth]{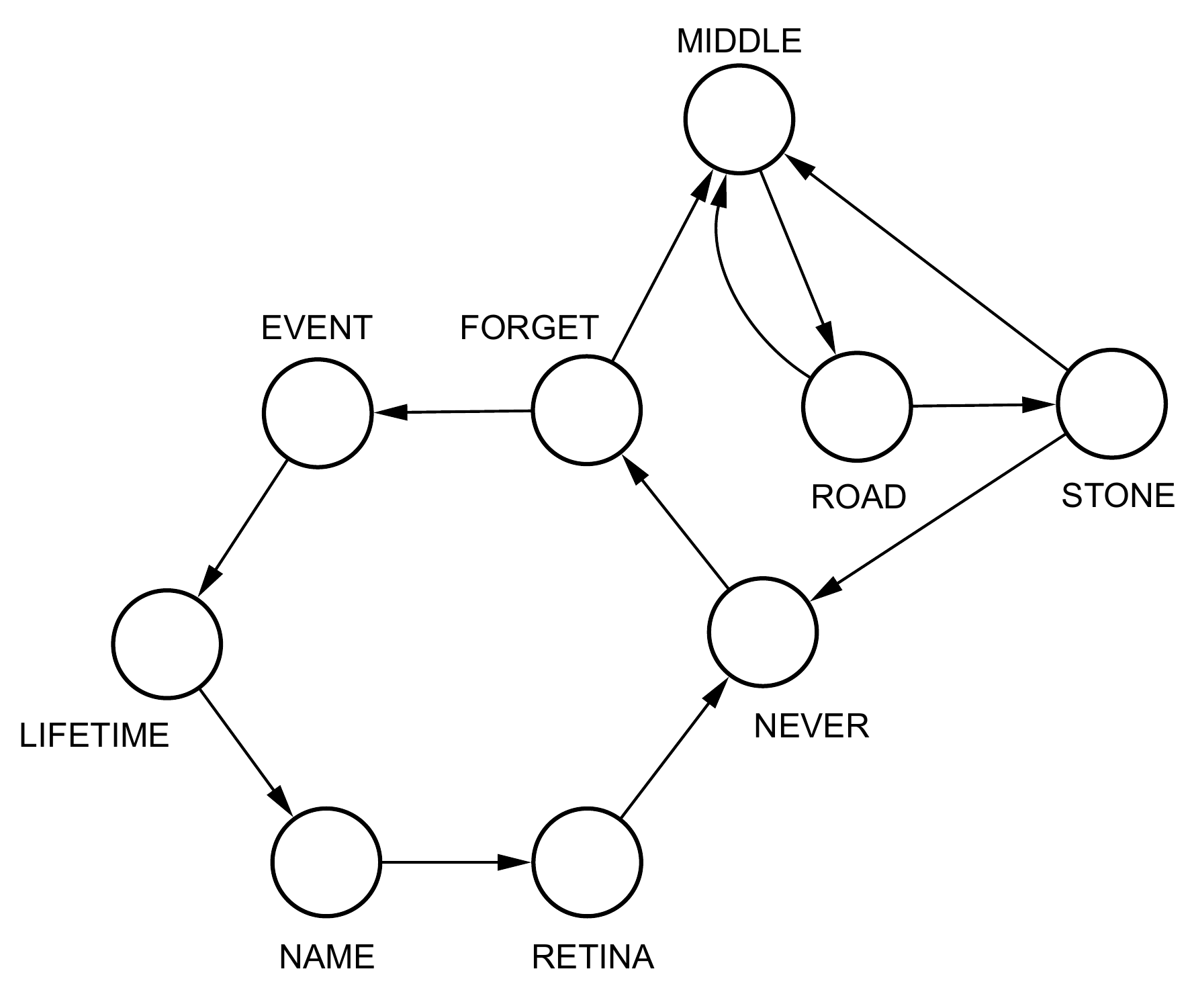}
  \caption{\label{fig.pre}Example of word adjacency network created from the poem ``In the Middle of the Road'', by Carlos Drummond de Andrade. The pre-processing steps performed to generate this word adjacency network are shown in table \ref{tab.pre}.}
\end{figure}

Before mapping the text into a word adjacency network, some pre-processsing steps are
usually applied. First, words conveying low semantic content, such as articles and prepositions, are removed from the text. These words, referred to as \emph{stopwords}, are disregarded because they just serve to link content words. As a consequence, they can be straightforwardly replaced by network edges. The remaining words are then lemmatized, i.e. they are transformed to their canonical forms. To assist the lemmatization process, all words are labeled with their part-of-speech tags. Particularly, in this study, I used a model based on maximum entropy~\cite{verref}.
To exemplify the construction of a word adjacency network, I show in table \ref{tab.pre} the pre-processing steps performed in a short text. The corresponding network is shown in figure \ref{fig.pre}.

\begin{table}[h]
\caption{\label{tab.pre}Pre-processing steps applied to the poem ``In the Middle of the Road'', by Carlos Drummond de Andrade. First, \emph{stopwords} are removed. Then, the remaining words are mapped to their canonical form via lemmatization. Note that transformations fatigued $\mapsto$ fatigue and retinas $\mapsto$ retina occur during the lemmatization step. The network obtained from the poem is illustrated in figure \ref{fig.pre}.}
\begin{tabular}{|c|l|}
\hline
{\bf Processing step} & {\bf Outcome} \\
\hline
\multirow{7}{*}{Original text}  & In the middle of the road there was a stone / there was a stone \\
              & in the middle of the road there was a stone in the middle of the\\
              & road there was a stone. Never should I forget this event / in the\\
              & lifetime of my fatigued retinas / Never should I forget that in \\
              & the middle of the road / there was a stone / there was a stone  \\
              & in the middle of the road / in the middle of the road there was \\
              & a stone. \\
\hline
                  & middle road stone stone middle road stone middle road stone never \\
Stopwords removal & forget event lifetime fatigued retinas never forget middle road stone \\
                  & stone middle road middle road stone \\
\hline
              & middle road stone stone middle road stone middle road stone never \\
Lemmatization & forget event lifetime fatigue retina never forget middle road stone \\
              & stone middle road middle road stone \\
\hline
\end{tabular}

\end{table}

\section{Topological characterization of networks} \label{asmedidas}

A network can be defined as $G=\{\mathbb{V},\mathbb{E}\}$, where $\mathbb{V}$ denotes the set of nodes and $\mathbb{E}$ denotes the set of edges, which serve to link nodes. An unweighted network can be represented by an adjacency matrix $\matr{A}=\{a_{ij}\}$, where each element $a_{ij}$ stores the information concerning the connectivity of nodes $i$ and $j$. If $i$ and $j$ are connected, then $a_{ij} = 1$. Otherwise, $a_{ij} = 0$. Note that, in a undirected network devoid of self loops, $\matr{A}^T = \matr{A}$.
Currently, there are several network measurements available to characterize the topology of complex networks~\cite{surveymeas}. Here, I describe the main measurements applied for text analysis:

\begin{itemize}

  \item {\bf Degrees}: the degree ($k$)  is the number of edges connected to the node, i.e. $k_i = \sum_j a_{ij}$.
      %
%
      A relevant feature related to the degree that has been useful for text analysis are the are the average node degree and the standard deviation of the neighbors, which are given by
      \begin{equation}
      \Red{
        k^{(n)}_{i} = \frac{1}{k_i} \sum_j a_{ij}  k_j,
      }
      \end{equation}
      \begin{equation}
      \Red{
        \Delta k^{(n)}_i = \Bigg{[} \frac{1}{k_i} \sum_j a_{ij} \Bigg{(} k_j - \frac{1}{k_i} \sum_m a_{im}  k_m  \Bigg{)}^ 2  \Bigg{]}^{1/2}.
      }
      \end{equation}
      In text networks, both $k^{(n)}$ and $\Delta k^{(n)}$ have been useful to quantify the structural organization of texts~\cite{58}.

  \item {\bf Accessibility}: the accessibility measurement is a extension of the node degree centrality~\cite{travencolo}. It is defined as a normalization of the diversity measurement~\cite{travencolo}, which quantifies the irregularity of a accessing neighbors through self-avoiding random walks~\cite{travencolo}. To define the accessibility, let $P_h(i,j)$ be the probability of a random walker starting at node $i$ to reach node $j$ in exactly $h$ steps. The heterogeneity of access to neighbors can be quantified with the diversity measurement:
      \begin{equation} \label{eq.div}
      \Red{
        {\delta}^{(h)}_i = -\sum_j P_h(i,j) \log P_h(i,j),
      }
      \end{equation}
      Given eq. \ref{eq.div}, the accessibility ($\alpha$) is computed as
      \begin{equation}
        \alpha^{(h)}_i = \exp ( \delta^{(h)}_i ).
      \end{equation}
      It can be shown that the accessibility is bounded according to the relation $0 \leq \alpha^{(h)} \leq n_h$, where $n_h$ is the number of nodes at the $h$-th concentric level~\cite{concentrico}. An example of the computation of the accessibility in a small network is shown in figure \ref{fig.acc}. In the example, nodes 2, 3, 4 and 5 belong to the first concentric level and nodes 6, 7, 8, 9 and 10 belong to the second concentric level. When one considers a regular access to the second level (red configuration), the accessibility corresponds to the total number of nodes located at the $h$-th concentric level. When some nodes are more accessed than others, the accessibility decreases because less nodes are effectively accessed.
      
      \begin{figure}
  \centering
  \includegraphics[width=0.7\textwidth]{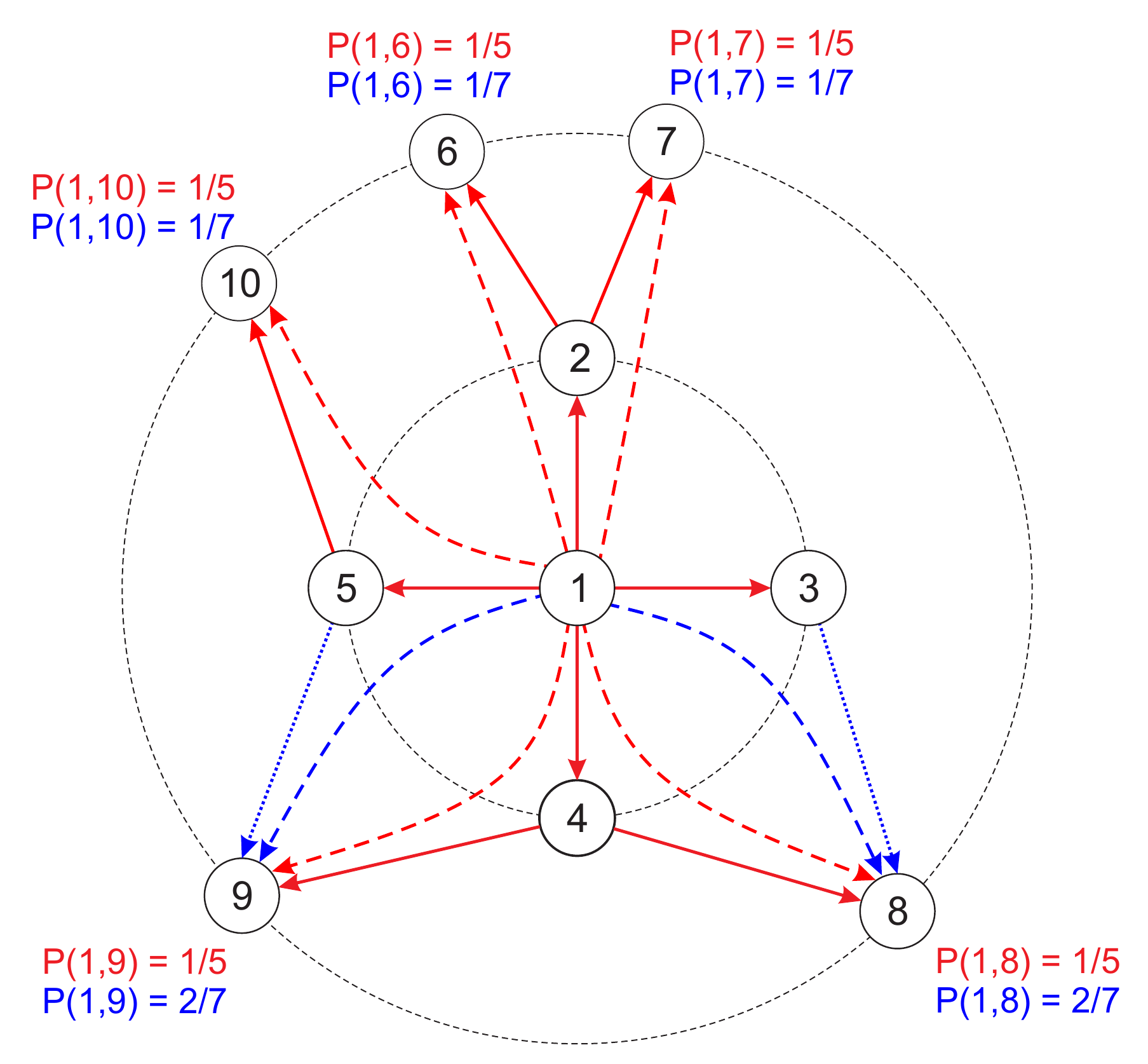} 
  \caption{\label{fig.acc}Probabilities of transition from node 1 considering $h=2$ steps for two distinct configuration of links. The first configuration considers only the red edges and the second one considers both red and blue edges. Note that, in the first configuration, the probability to reach any node at the second level is the same. In this case, $\alpha_1^{(h=2)} = 5$. When blue edges are included, nodes 7 and 9 tend to receive more visits than the other nodes, according to the considered probabilities. For this reason, the effective number of accessed nodes drops to $\alpha_1^{(h=2)} = 4.71$.}
\end{figure}

      In textual networks, it has been shown that the accessibility is more advantageous than other traditional centrality measurements as it is able to capture more information at further hierarchical levels~\cite{concentrico}. This measurement has been successfully applied to detect core concepts in texts~\cite{extractive}. Furthermore, it has also been employed to generate informative summaries~\cite{extractive}. A dependency of the distribution of this measurement with stylistic features of text was observed in~\cite{probing}.

  \item {\bf Betweenness}: the betweenness is a centrality measurement that considers that  a node is relevant if it is highly accessed via shortest paths. If $n_{sit}$ is the number of geodesic paths between $s$ and $t$ passing through node $i$; and  $n_{st}$ is the total number of shortest paths linking $s$ and $t$, then the betweenness is defined as:
      \begin{equation}
        B_i = \frac{1}{N^2} \sum_s \sum_t \frac{n_{sit}}{n_{st}}
      \end{equation}
      In word adjacency networks, high frequent words usually take high values of betweenness. However, some words may act as articulation points whenever they link two semantical contexts or communities~\cite{commrev}. It has been shown that the betweenness is able to identify the generality of contexts in which a word appears~\cite{comparing}. More specifically, domain-specific words tend to assume lower values of betweenness when compared with more generic words.

  \item {\bf Assortativity}: several real-world networks are formed of nodes with a specific type of classification. For example, in social networks, individuals may be classified by considering their age, sex or race. When analyzing the connectivity patterns of networks, it might be relevant to study how distinct classes connect to each other. This type of analysis is usually performed with the so-called assortativity measurement~\cite{mixing}.
      %
      %
      %
      \Red{The assortativity can be computed as the Pearson correlation coefficient} ($r$):
      \begin{equation}
            r = \frac{e^{-1}\sum_{j>i}^{ } k_{i}k_{j}a_{ij} - \big[ e^{-1}\sum_{j>i}^{ } \frac{1}{2}(k_{i}+ k_{j})a_{ij}\big]^{2} }{e^{-1} \sum_{j>i}^{ } \frac{1}{2}(k_{i}^{2}+ k_{j}^{2})a_{ij}- \big[ e^{-1}\sum_{j>i}^{ } \frac{1}{2}(k_{i}+ k_{j})a_{ij}\big]^{2}},
      \end{equation}
      where $e$ is the total number of edges. In word adjacency networks, the assortativity quantifies how words with distinct frequency appear as neighbors.

  \item {\bf Clustering coefficient}: the clustering coefficient ($C$) quantifies the local density of neighbors of a given node. The local definition of the clustering coefficient is given by the fraction of the number of triangles among all possible connected sets of three nodes:
      \begin{equation}
        C = 3 \sum_{k>j>i} a_{ij} a_{ik} a_{jk} \Bigg{[} \sum_{k>j>i} a_{ij} a_{ik} + a_{ji} a_{jk} + a_{ki} a_{kj} \Bigg{]} ^{-1}.
      \end{equation}
      Similarly to the betweenness, the clustering coefficient is useful to detect words appearing in generic contexts~\cite{comparing}. However, differently from the betweenness, the clustering coefficient analyzes only the local neighborhood of nodes.

  \item {\bf Average shortest path length}: the average shortest path length ($l$) is the typical distance between any two nodes in the network. \Red{This measurement was used in this paper because it has been useful in stylistic-based applications}~\cite{comparing}. In texts, the average shortest path length quantifies words relevance. More specifically, according to this measurement, the most relevant words are those that are close to the hubs.

\end{itemize}

Most of the measurements described in this section are local measurements, i.e. each node $i$ possesses a value $\tilde{X}_i$, where $\tilde{X}=\{k,k^{(n)},\Delta k^{(n)},\alpha,B,C,l\}$. For the purposes of this paper, \Red{it is necessary to sum up the local measurements}. The most natural choice is to characterize the documents by computing the average $\langle \tilde{X} \rangle$, where $\langle \ldots \rangle = N^{-1} \sum_{i=1}^N \ldots$ stands for the average computed over the $N$ distinct words of the text. A disadvantage associated with this type of \Red{summing procedure} is that all words receive the same weight, regardless of their number of occurrences in the text. To avoid this potential problem, I also computed the average value $\langle \ldots \rangle^* = \eta^{-1} \sum_{i=1}^\eta \ldots$ obtained when only the $\eta=50$ most frequent words are considered. The standard deviations $\Delta \tilde{X}$ and the skewness $\gamma(\tilde{X})$ were also used to characterize the documents.

A drawback associated to the computation of local topological measurements in word adjacency networks is the high correlation found between these measurements and the node degree (i.e. the word frequency). To minimize this correlation, the following procedure was adopted. Each of the measurements was normalized by the average obtained over $30$ texts produced using a word shuffling technique, where the frequencies of words are preserved. If $\mu(\tilde{X}^{(R)})$ and $\sigma(\tilde{X}^{(R)})$ are the average and deviation obtained over the random realizations, the normalized measurement $X$ and the error $\epsilon(X)$ related to $X$ are
\begin{equation}
    X = \frac{\tilde{X}}{\mu(\tilde{X}^{(R)})}
\end{equation}
\begin{equation}
    \Red{
    \epsilon(X) = \frac{\sigma(\tilde{X}^{(R)})}{\mu(\tilde{X}^{(R)})^2} \tilde{X} =  \frac{\sigma(\tilde{X}^{(R)})}{\mu(\tilde{X}^{(R)})} X.
    }
\end{equation}

\section{Traditional stylistic features} \label{tradicao}

\subsection*{Frequency of words and characters}

Traditional methods usually perform statistical analysis using specific textual features~\cite{manning}. An important contribution to the stylometry was introduced by Mosteller and Wallace~\cite{mosteller}, which showed that the frequency of function words (such as \emph{any}, \emph{of}, \emph{a} and \emph{on}) is useful to characterize the style in texts. Frequent words have also been used in strategies devised by physicists where the distance of frequency ranks was used to compute the similarity between texts~\cite{28,29,30}. Another relevant feature for characterizing styles in texts is the frequency of character bigrams \Red{(i.e. a sequence of two adjacent characters)}. These attributes have proven useful e.g. to detect the stylistic marks of specific authors~\cite{52}.

\subsection*{Intermittence}

\begin{figure}
  \centering
  \includegraphics[width=0.9\textwidth]{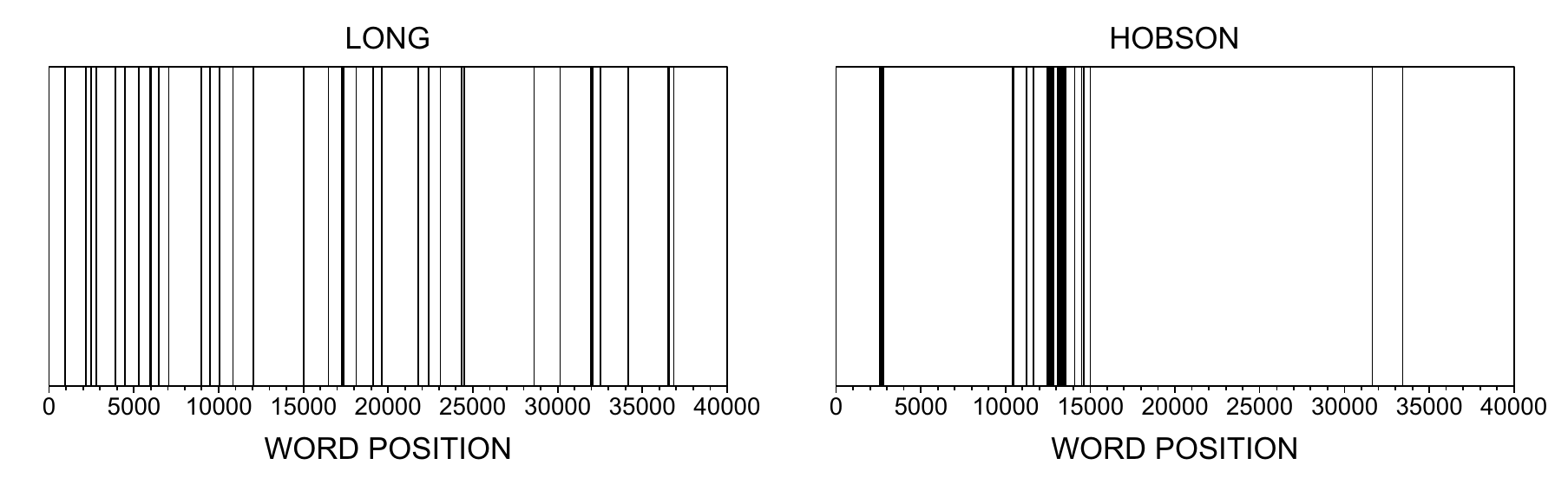}
  \caption{\label{figintt}Profile of the spatial distribution of \emph{long} ($N_i = 44$ and $I_i = 1.02$) and \emph{Hobson} ($N_i = 45$ and $I_i = 3.40$) in the book ``Adventures of Sally'', by P. G. Wodehouse. Because \emph{Hobson} is unevenly distributed along the book, this word take high values of intermittency.
  }
\end{figure}

The uneven spatial distribution of words along texts is a feature useful for characterizing the style of texts~\cite{statistical}. The quantification of the homogeneity of the distribution of words along texts can be performed by using recurrence times, a standard measure employed to study time series~\cite{rtimes}.
In texts, the concept of time is represented in terms of the number of words occuring in a given interval. For each word $i$, the recurrence time  $T_j$ is defined as the number of words appearing between two successive occurrences of $i$ plus one. For example, the recurrence times of the word ``stone'' in the lemmatized text shown in table \ref{tab.pre} are $T_1 = 1$, $T_2 = 3$, $T_3 = 3$, $T_4 = 11$, $T_5 = 1$ and $T_6 = 5$.
If a word $i$ occurs $N_i$ times in a text comprising
$N_T$ words, it generates a sequence of $N_i-1$ recurrence times $\{T_1, T_2,\ldots,T_{N_i-1} \}$. In order to consider the time $T_f$ until the first occurrence of $i$ and the time $T_l$, the number of words between the last occurrence of $i$ and the last word of the text, the recurrence time $T_N = T_f + T_l$ is added to the set of recurrence times of word $i$. As such, $\langle T \rangle = N_T / N_i$, where $\langle \cdot \rangle$ is the average over distinct $T_j$'s. Note that the average recurrence time $\langle T \rangle$ does not provide additional information, since it only depends on the frequency. The intermittence (or burstiness) of the word $i$ is obtained from the coefficient of the variation of the recurrence times:
\begin{equation}
    I = \sigma_T / \langle T \rangle = \Bigg{[} \frac{\langle T^2 \rangle}{\langle T \rangle^2} - 1 \Bigg{]}^{1/2}.
\end{equation}
It has been shown that the intermittence has been useful to identify core concepts in texts, even when a large corpus is not available. Relevant words usually take high values of intermittency, i.e. $I \gg 1$. \emph{Stopwords}, on the other hand, are evenly distributed along the text~\cite{probing}. To illustrate the properties of the intermittency measurement, figure \ref{figintt} shows the spatial distribution of the words \emph{long} and \emph{Hobson} in the book ``Adventures of Sally'', by P.G. Wodehouse. The frequency of these words are similar, yet their values of intermittency are quite different. Note that, while the distribution of \emph{long} is relatively homogeneous, \emph{Hobson} is unevenly distributed along the text. Because bursty concepts are the most relevant words~\cite{statistical}, in the example of figure~\ref{figintt}, \emph{Hobson} would be considered a relevant character in the plot.

\section{Supervised classification}

In this section, I describe the two methods employed to combine distinct strategies of classification. The objective here is to combine evidences from both traditional and CN-based methods. Before introducing the methods for combining classifiers, the main concepts concerning the supervised learning task are presented.

In a typical supervised learning task, two datasets are employed: the training and the test dataset. The training dataset $X_{\textrm{tr}} = \{(x_1,y_1),(x_2,y_2),\ldots,(x_l,y_l)\}$ is the set of instances whose classes are known beforehand. The first component of the tuples $x_i = (f_1 = a_1, f_2 = a_2,\ldots)$ represents the values of attributes used to describe the $i$-th instance.
The second element $y_i \in \mathcal{Y} = \{y_1,y_2,\ldots\}$ represents the class label of the $i$-th training instance. In the supervised learning task, the objective is to obtain the map $x \mapsto y$. The quality of the map obtained is evaluated with the test dataset $X_{\textrm{ts}} = \{(x_l,y_l),(x_{l+1},y_{l+1})\ldots,(x_{l+u},y_{l+u})\}$. The technique used to evaluate the performance of the classifiers used in this study is the well-known 10-fold cross-validation~\cite{wekabook}.

\Red{
An usual procedure in classification tasks is the quantification of the relevance of each attribute for the task. To quantify the relevance of attributes to discriminate the data, several indices have been proposed~\cite{wekabook}. A well-known index, the information gain, quantifies the homogeneity of the set of instances in $X_{\textrm{tr}}$ when the value of the attribute is specified~\cite{infogain}. Whenever a single class prevails when the value of an attribute $f_k$ is specified, the information gain associated to $f_k$ takes high values.
Mathematically, the information gain $\Omega$ is defined as
\begin{equation}
    \Omega(X_{\textrm{tr}}, f_k) = \mathcal{H}(X_{\textrm{tr}}) - \mathcal{H}(X_{\textrm{tr}}|f_k)
\end{equation}
where $\mathcal{H}(X_{\textrm{tr}})$ is the entropy of the training dataset $X_{\textrm{tr}}$ and $\mathcal{H}(X_{\textrm{tr}}|f_k)$ is the entropy of the training dataset when the $k$-th attribute is specified. The quantity $\mathcal{H}(X_{\textrm{tr}}|f_k)$ is computed as
\begin{equation}
    \mathcal{H}(X_{\textrm{tr}}|f_k) = \frac{1}{{|X_{\textrm{tr}}|}} \sum_{v \in V(f_k)} {|\{ x \in X_{\textrm{tr}} | f_k = v \} |~ }  \mathcal{H}(\{ x \in X_{\textrm{tr}} | f_k = v \})
\end{equation}
where $V(F_k)$ is the set comprising all values taken by $f_k$ in $X_{\textrm{tr}}$.
}

\subsection*{Hybrid classifier}

To define this classifier, consider the following definitions. \Red{Let $m_{ij}$ be the strength associating the $i$-th instance to the $j$-th class, where $0 \leq m_{ij} \leq 1$. The term $m_{ij}$, henceforth referred to as \emph{membership strength}, can be interpret as the likelihood of instance $i$ to belong to class $j$. Note that the quantification of $m_{ij}$ depends on the classifier being used. A detailed description of methods for computing the membership strength can be found in~\cite{fuzzyknn,fsvm,ftree,fforest,fmlp}.}
Let $m^{(R)}_{ij}$ be the membership strength obtained when topological features of complex networks are used and $m^{(T)}_{ij}$ the membership strength obtained when traditional statistical features are analyzed. According to the hybrid strategy, the combination of both topological and traditional methods is achieved according to the following convex combination
\begin{equation} \label{eq.combine}
    m^{(H)}_{ij} = \lambda m^{(R)}_{ij} + ( 1 - \lambda ) m^{(T)}_{ij},
\end{equation}
where $\lambda \in [0,1]$ accounts for the weight associated to the topological strategy. Note that the combination of evidences performed in equation \ref{eq.combine} yields a membership strength $m^{(H)}_{ij}$ ranging in the interval $[0,1]$. The final decision is then made according to the rule
\begin{equation}
    \Red{
    \tilde{y}_i = \max_j m^{(H)}_{ij},
    }
\end{equation}
where $\tilde{y}_i \in \mathcal{Y}$  denotes the correct class label associated to the $i$-th test instance.
\Red{
In practical applications, a screening on the $\lambda$ can be useful to find its best value. However, this process might be computationally unaffordable for very large datasets. In this case, the process of finding the adequate value of the parameter can be made via application of optimization heuristics}~\cite{grid1,grid2}.

\begin{figure}
  \centering
  \includegraphics[width=1\textwidth]{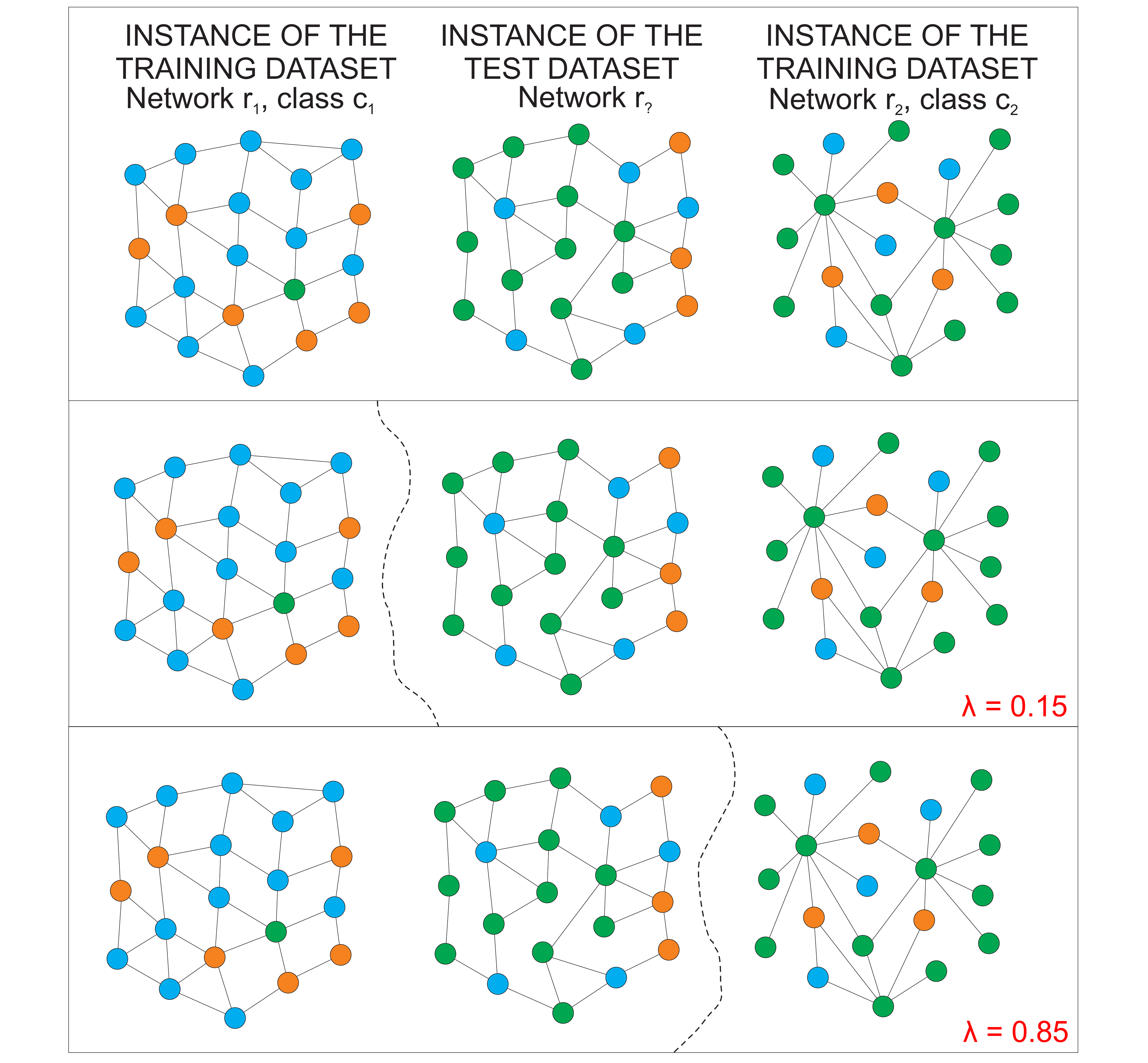}
  \caption{\label{fig.clhyb}Example of classification based on the hybrid classifier. In the top panel, the network $r_?$ might assume two possible classes: $c_1$ and $c_2$. An example for each of these classes is provided (see networks $r_1$ and $r_2$). In the central panel, the decision boundary obtained for $\lambda = 0.15$ is shown. Because $\lambda$ takes a low value in this case, the decision is mainly based on the number of shared nodes (words). As a consequence, $r_?$ is classified as belonging to class $c_2$. For higher values of $\lambda$, the topological features of texts takes over. In the bottom panel, $r_?$ is classified as belonging to class $c_1$ because $r_?$ and $r_1$ are topologically similar.
  }
\end{figure}

To illustrate how the decision is performed with the hybrid algorithm, figure \ref{fig.clhyb} shows the classification of a text modeled as a network.
In the top panel, the central network is the text whose class is to be inferred. The left and right networks represent two networks in the training dataset. \Red{The node color} denotes the node label, i.e. the word associated to the node. In this example, two texts are semantically similar if they share the same words.
In the central panel, the scenario where the hybrid classified is set with $\lambda = 0.15$ is considered. Because $\lambda$ is close to zero, the classification is mostly based on the number of shared words. In fact, the decision boundary in this case is created so that the test instance is classified in the same class as the right training instance.
The bottom panel illustrates the decision made with $\lambda=0.85$. Now, the topological features of the text is the most prevalent feature used for the classification. As a consequence, the boundary decision moves to the right side so that the test instance is classified as the same class as the left training instance. Note that, the test instance as belonging to class $c_1$ because the $r_1$ and $r_?$ are topologically similar.

\subsection*{Tiebreaker classifier}

\begin{figure}
  \centering
  \includegraphics[width=0.75\textwidth]{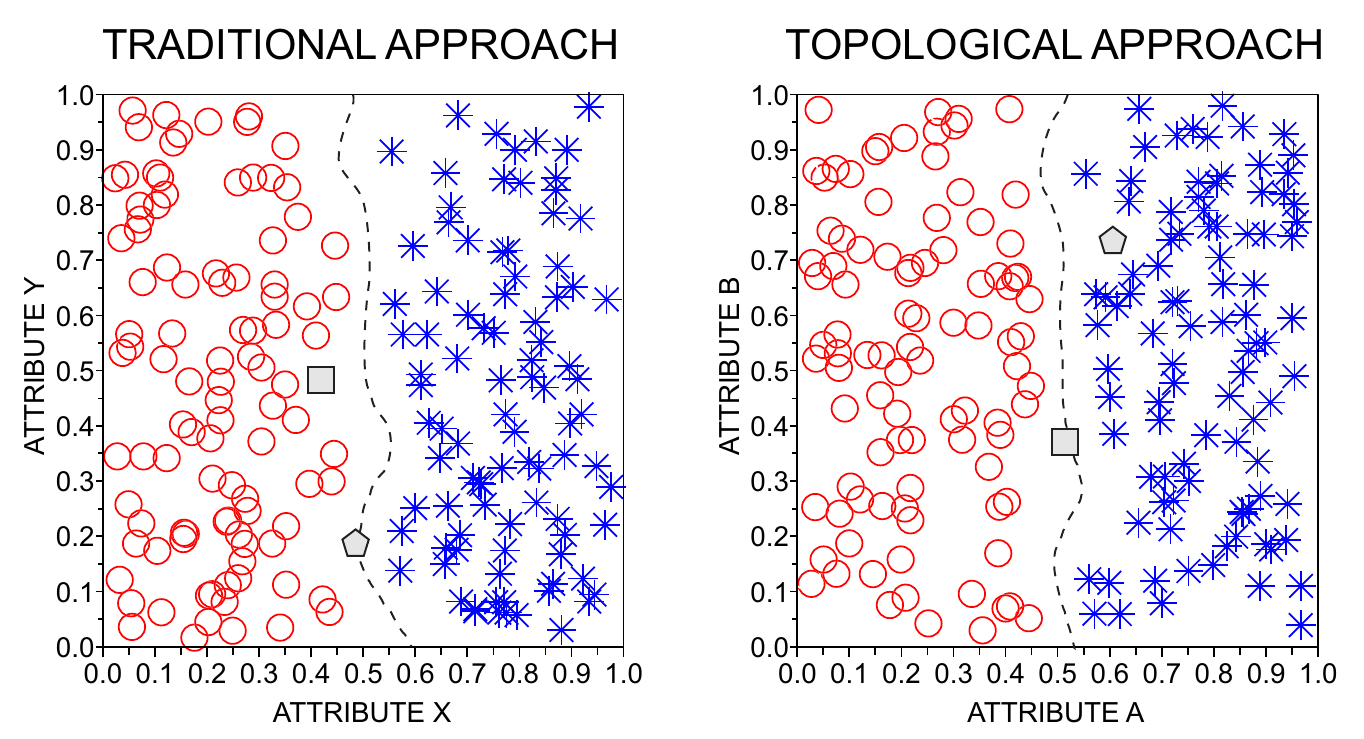}
  \caption{\label{theta}Example of classification based on the tiebreaker classifier. In the left panel, the gray instances are the test instances that should be classified with the class labels red circle or blue asterisk. In this case, traditional attributes were used. Note that the square is significantly far from the decision boundary (dashed line). Therefore, the classification of this instance \Red{does not demand the use of topological features} because $\Delta \geq \theta$ in equation \ref{lab.tiebreaker}. Differently, the pentagon is located on the decision boundary. Because $\Delta < \theta$ in this case, topological attributes are used to perform the classification. According to the topological attributes (see right panel), the test instance represent by the pentagon is classified as belonging to the blue class.
  }
\end{figure}

Similarly to the hybrid classifier, this classification scheme uses as attributes both traditional and topological features of texts modelled as networks. The objective of this approach is to use the topological attributes only when the classification performed with traditional features is not reliable, as revealed by the values of membership strength. Consider that the two most likely classes to which the unknown instance belongs are $j$ and $k$, according to traditional features. As a consequence, $m^{(T)}_{ij} \geq m^{(T)}_{ik} \geq m^{(T)}_{il}$, for each class $l \in \mathcal{Y} - \{i,j\}$. Let $\Delta$ be the difference between the probabilities associated to the two most likely classes, i.e. $\Delta = m^{(T)}_{ij} - m^{(T)}_{ik}$. If such difference surpasses a given threshold $\theta$, the tiebreaker performs the classification by using only traditional attributes. Conversely, topological attributes are employed to infer the class of the unknown instance. Equivalently,
\begin{equation} \label{lab.tiebreaker}
    \tilde{y}_i =\left\{
    \begin{array}{l l}
      \max_j m^{(T)}_{ij}, & \textrm{if $\Delta = m^{(T)}_{ij} - m^{(T)}_{ik} \geq \theta$} \\
      \max \{ m^{(R)}_{ij}, m^{(R)}_{ik} \},  & \textrm{otherwise.}
\end{array}\right.
\end{equation}
Note that the threshold $\theta$ ultimately decides which attributes (traditional or topological) are used to perform the classification. An example of classification using this approach is shown in figure \ref{theta}.

\section{Results and discussion} \label{osresultados}

The effectiveness of the combination of traditional textual features and topological measurements of networks was evaluated in the context of two natural language processing tasks. Both tasks rely upon the characterization of stylistic marks of texts. The studied tasks were the authorship attribution and genre detection problems. The classifiers chosen to compound the combining techniques were: kNN, Support Vector Machines (SVM), Random Forest (RFO) and Multilayer Perceptron (MLP). These classifiers were chosen because they usually yield good accuracy rates with default parameters~\cite{systematic}.

\subsection{Authorship recognition} \label{aut.rec}

\begin{figure}
  \centering
  \includegraphics[width=1\textwidth]{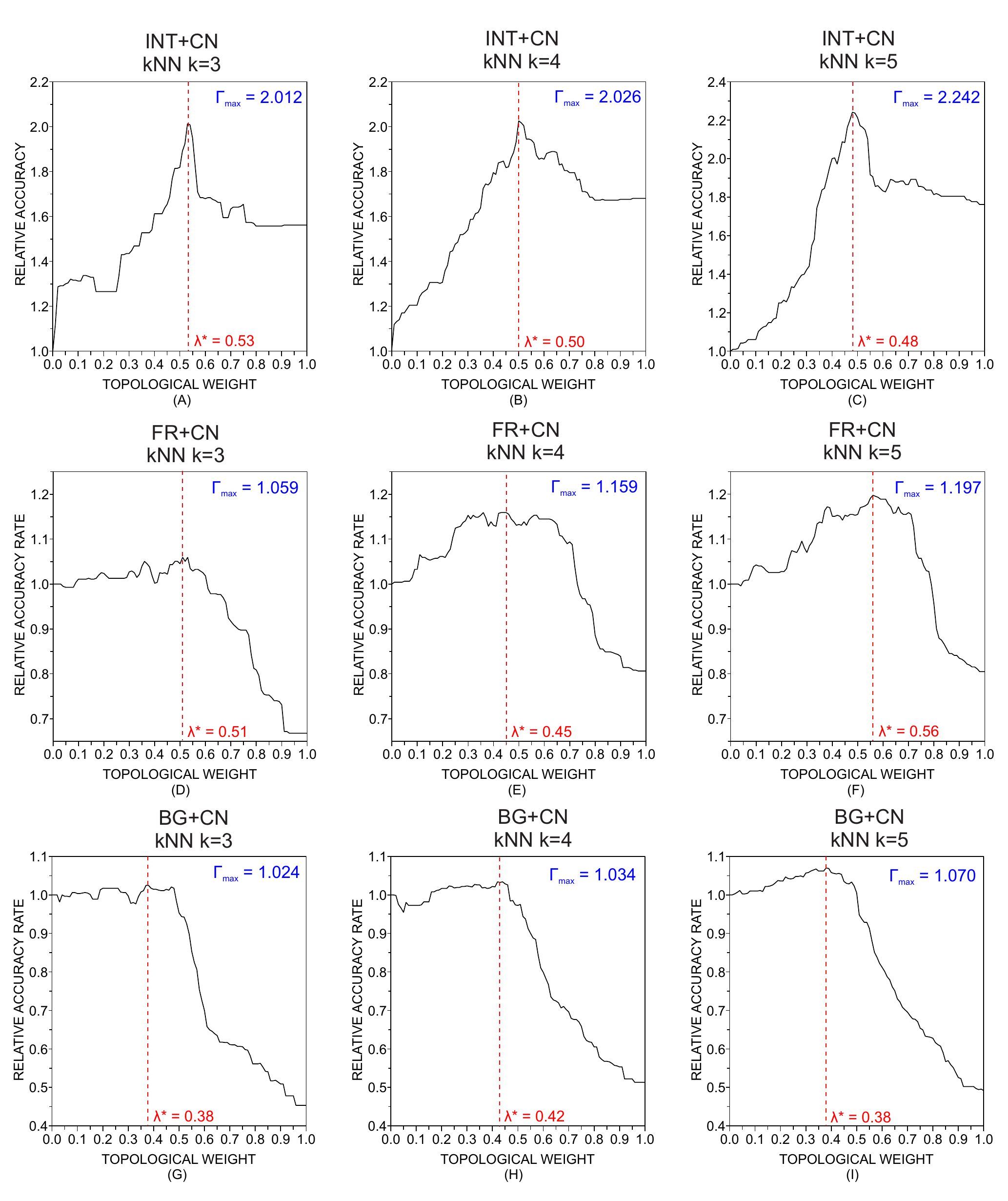}
  \caption{\label{fig.lambda.aut}Relative accuracy rate as a function of the topological weight ($\lambda$) obtained with the k nearest neighbors. The attributes employed were: (a)-(c): intermittence; (d)-(f): stopwords; (g)-(i): characters. The parameters $k$ of the $k$-nearest neighbors were: $k=3$ in (a), (d) and (g); $k=4$ in (b), (e) and (h); and $k=5$ in (g), (f) and (i). As it turns, there is an improvement of the accuracy rates when traditional methods are combined with the technique based on the topological analysis of complex networks.}
\end{figure}

In the authorship attribution task, the objective is to identify the authors of texts whose identity is lacking~\cite{t5}. The dataset employed here comprises books written by eight authors: Arthur Conan Doyle (ACD), Bram Stoker (BRS), Charles Dickens (CHD), Thomas Hardy (THH), Pelham Grenville Wodehouse (PGW), Hector Hugh Munro (HHM) and Herman Melville (HME). To show how the topological properties of the networks modelling books can be useful to improve the characterization of authors' styles in texts, the following combination of attributes were considered:
\begin{itemize}

  \item {\bf INT+CN}: the intermittence of stopwords were considered along with the topological measurements of the network modeling the text. As stopwords, I considered all the words that
      appeared at least once in all books of the dataset. \Red{According to the dataset, there is a total of $340$ stopwords.}

  \item {\bf FR+CN}: the frequency of stopwords were considered along with the topological attributes of the networks.

  \item {\bf BG+CN}: the frequency of character bigrams were considered along with the topological attributes of the networks. All the possibilities of character bigrams ($634$) were considered in this analysis.

\end{itemize}
For each combination of attributes, the gain in performance when the topology is considered as an additional feature is represented as:
\begin{equation}
    \Delta\Gamma(x) = \frac{\Gamma_H(x) - \Gamma_T}{\Gamma_T},
\end{equation}
where $\Gamma_H(x)$ denotes the accuracy rate obtained with either the hybrid or tiebreaker classifier and $x$ is the corresponding parameter employed, i.e. $\lambda$ for the hybrid classifier and $\theta$ for the tiebreaker classifier. Figure \ref{fig.lambda.aut} shows the values of  $\Delta\Gamma(\lambda)$ when the kNN is used to compose the hybrid classifier. The results obtained for additional values of the parameter $k$ are shown in table \ref{tab.lambda.aut}. Figure \ref{fig.lambda.aut} also shows the maximum accuracy rate obtained with the variation of $\lambda$:
\begin{equation}
    \Delta\Gamma_{max} = \max_{\lambda \in [0,1]} \Delta\Gamma(\lambda).
\end{equation}
With regard to the INT+CN combination, the addition of topological features of texts yielded accuracy rates much higher than the ones obtained with intermittency features alone, since $\Delta\Gamma_{max} > 2$ for $k=\{3,4,5\}$. Note that, in this case, the network approach alone performed better than the method based solely on intermittency features, because $\Delta\Gamma(\lambda=1) > 1$. This observation might explain the high gain obtained when networks are included as an additional feature.
As for the FR+CN combination, the highest gain in performance was $\Delta\Gamma_{max} = 1.197$, which was obtained for $k=5$. The lowest gain in accuracy occurred for the BIG+CN combination. In this case, the maximum gain was $\Delta\Gamma_{max} = 1.070$ for $k=5$.
When other classifiers were used to compose the hybrid classifier, similar results have been found (see table \ref{tab.lambda.aut.o}): the highest and lowest improvement in performance occurred for the INT+CN and BIG+CN combinations, respectively. In general, the hybrid classification that combined traditional statistical features and topological measurements of networks improved the classification performance in the authorship recognition task. Interestingly, in several cases, the combination outperformed the accuracy rates obtained when the compounding strategies were analyzes separately, i.e.
$\Gamma_H(\lambda \in ]0,1[) > \Gamma_H(\lambda = 0)$ and $\Gamma_H(\lambda \in ]0,1[) > \Gamma_H(\lambda = 1)$.

\begin{table}[h]
\centering
\caption{\label{tab.lambda.aut}Best relative accuracy rate $\Delta\Gamma_{max}$ obtained with the hybrid classifier based on the k-nearest neighbors method. The threshold employed $\lambda^*$ to obtain the best accuracy rate is also shown. Note that the improvement of accuracy occurred in several occasions.}
\begin{tabular}{|l|cc|cc|cc|cc|cc|}
\hline
 & \multicolumn{2}{|c|}{$k=1$} & \multicolumn{2}{|c|}{$k=2$} & \multicolumn{2}{|c|}{$k=3$} &\multicolumn{2}{|c|}{$k=4$} & \multicolumn{2}{|c|}{$k=5$} \\
 \cline{2-11}
 & $\Delta\Gamma_{max}$ & $\Delta\lambda^*$ & $\Delta\Gamma_{max}$ & $\lambda^*$ & $\Delta\Gamma_{max}$ & $\lambda^*$ & $\Delta\Gamma_{max}$ & $\lambda^*$ & $\Delta\Gamma_{max}$ & $\lambda^*$ \\
\hline
{\bf Intermittence} & 1.252 & 0.55 & 1.524 & 0.57 & 2.012 & 0.53 & 2.026 & 0.50 & 2.242 & 0.48 \\
{\bf Stopwords}     & 1.000 & 0.00 & 1.097 & 0.41 & 1.059 & 0.51 & 1.159 & 0.45 & 1.197 & 0.56 \\
{\bf Characters}    & 1.000 & 0.00 & 1.000 & 0.00 & 1.024 & 0.38 & 1.034 & 0.44 & 1.070 & 0.37 \\
\hline
\end{tabular}
{}
\end{table}

\begin{table}[h]
\centering
\caption{\label{tab.lambda.aut.o}Best relative accuracy rate $\Delta\Gamma_{max}$ obtained with the hybrid classifier based on the Support Vector Machine (SVM), Random Forest (RFO) and Multi Layer Perceptron (MLP) methods. The threshold employed $\lambda^*$ to obtain the best accuracy rate is also shown. Note that the improvement of accuracy occurred in several occasions.}
\begin{tabular}{|l|cc|cc|cc|}
\hline
 & \multicolumn{2}{|c|}{{\bf SVM}} & \multicolumn{2}{|c|}{\bf RFO} & \multicolumn{2}{|c|}{\bf MLP} \\
 \cline{2-7}
 & $\Delta\Gamma_{max}$ & $\lambda^*$ & $\Delta\Gamma_{max}$ & $\lambda^*$ & $\Delta\Gamma_{max}$ & $\lambda^*$ \\
\hline
{\bf Intermittence} & 1.125 & 0.25 & 1.800 & 0.59 & 1.227 & 0.53 \\
{\bf Stopwords}     & 1.176 & 0.49 & 1.529 & 0.36 & 1.045 & 0.50 \\
{\bf Characters}    & 1.032 & 0.20 & 1.000 & 0.00 & 1.000 & 0.00 \\
\hline
\end{tabular}
{
}
\end{table}

\begin{figure}
  \centering
  \includegraphics[width=1\textwidth]{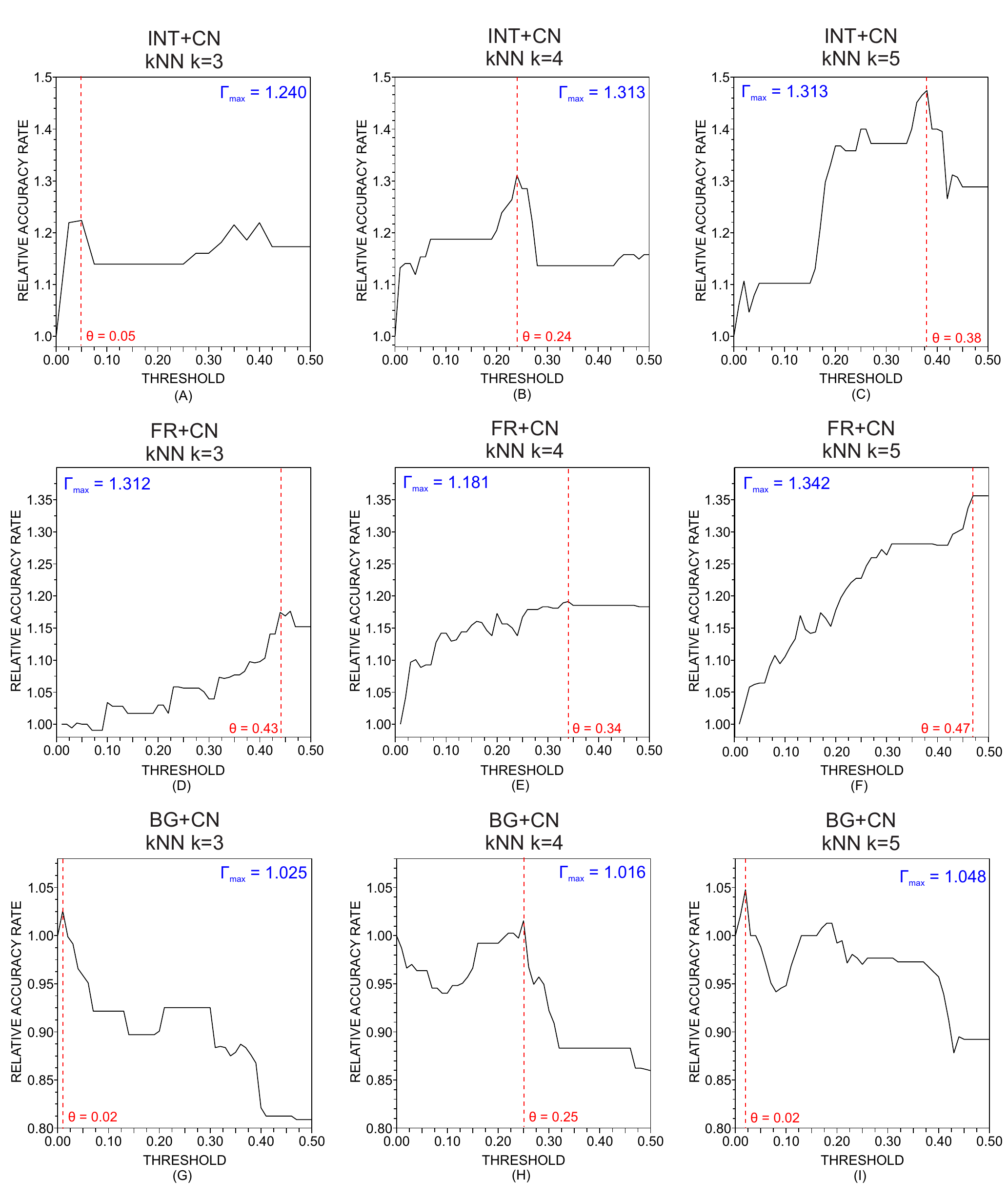}
  \caption{\label{fig.asw}\Red{Relative accuracy rate as a function of the threshold ($\theta$) obtained with the tiebreaker algorithm applied to the the k nearest neighbors.} The attributes employed were: (a)-(c): intermittence; (d)-(f): stopwords; (g)-(i): characters. The parameters $k$ of the $k$-nearest neighbors were: $k=3$ in (a), (d) and (g); $k=4$ in (b), (e) and (h); and $k=5$ in (g), (f) and (i). As it turns, there is an improvement of the accuracy rates when traditional methods are combined with the technique based on the topological analysis of complex networks.}
\end{figure}

Traditional and topological attributes of texts were also combined using the tiebreaker classifier. The results obtained with the kNN classifier for $k=\{3,4,5\}$ are shown in figure \ref{fig.asw}. The results obtained for other values of the parameter $k$ are displayed in table \ref{tab.tie.knn.aut}. The tiebreaker classifier built using the INT+CN combination provided an increase in performance of up to 31\%. The maximum gain, obtained with the combinations FR+CN and BG+CN, were respectively 34.2\% and 4.8\%. With regard to the other classifiers compounding the tiebreaker technique, the results obtained are shown in table \ref{tab.tie.out}. Interestingly, there was no improvement when the SVM was used ($\Delta\Gamma_{max}=1$). Differently, when the RFO classifier was used, an improvement of performance was observed in all three combinations of attributes. Finally, the tiebreaker classifier built with the MLP only provided a gain in accuracy for the combination INT+CN. While providing a gain in performance in several scenarios, the tiebreaker technique usually performed worst than the hybrid classifications performed with the same compound classifiers and parameters.

\begin{table}[h]
\centering
\caption{\label{tab.tie.knn.aut}Best relative accuracy rate $\Delta\Gamma_{max}$ obtained with the tiebreaker method based on the k-nearest neighbors algorithm. \Red{The threshold employed $\theta^*$ to obtain the best accuracy rate is also shown.} Note that the improvement of accuracy occurs in several occasions.}
\begin{tabular}{|l|cc|cc|cc|cc|cc|}
\hline
 & \multicolumn{2}{|c|}{$k=1$} & \multicolumn{2}{|c|}{$k=2$} & \multicolumn{2}{|c|}{$k=3$} &\multicolumn{2}{|c|}{$k=4$} & \multicolumn{2}{|c|}{$k=5$} \\
 \cline{2-11}
 & $\Delta\Gamma_{max}$ & $\theta^*$ & $\Delta\Gamma_{max}$ & $\theta^*$ & $\Delta\Gamma_{max}$ & $\theta^*$ & $\Delta\Gamma_{max}$ & $\theta^*$ & $\Delta\Gamma_{max}$ & $\theta^*$ \\
\hline
{\bf Intermittence} & 1.000 & -- & 1.112 & 0.01 & 1.241 & 0.31 & 1.311 & 0.24 & 1.474 & 0.38 \\
{\bf Stopwords}     & 1.000 & -- & 1.056 & 0.31 & 1.131 & 0.46 & 1.181 & 0.34 & 1.341 & 0.47 \\
{\bf Characters}    & 1.000 & -- & 1.027 & 0.02 & 1.026 & 0.01 & 1.016 & 0.25 & 1.048 & 0.02 \\
\hline
\end{tabular}
{}
\end{table}

\begin{table}[h]
\centering
\caption{\label{tab.tie.out}Best relative accuracy rate $\Delta\Gamma_{max}$ obtained with the tiebreaker algorithm applied to the Support Vector Machine (SVM), Random Forest (RFO) and Multi Layer Perceptron (MLP) methods. \Red{The threshold employed $\theta^*$ to obtain the best accuracy rate is also shown.} Note that the improvement of accuracy depends upon the pattern recognition method and attributes employed.. }
\begin{tabular}{|l|cc|cc|cc|}
\hline
  & \multicolumn{2}{|c|}{{\bf SVM}} & \multicolumn{2}{|c|}{\bf RFO} & \multicolumn{2}{|c|}{\bf MLP} \\
  \cline{2-7}
  & $\Delta\Gamma_{max}$ & $\theta^*$ & $\Delta\Gamma_{max}$ & $\theta^*$ & $\Delta\Gamma_{max}$ & $\theta^*$ \\
\hline
{\bf Intermittence} & 1.000 & -- & 1.375 & 0.02 & 1.055 & 0.02 \\
{\bf Stopwords}     & 1.000 & -- & 1.294 & 0.23 & 1.000 & -- \\
{\bf Characters}    & 1.000 & -- & 1.030 & 0.06 & 1.000 & -- \\
\hline
\end{tabular}
{}
\end{table}

\begin{figure}
  \centering
  \includegraphics[width=0.75\textwidth]{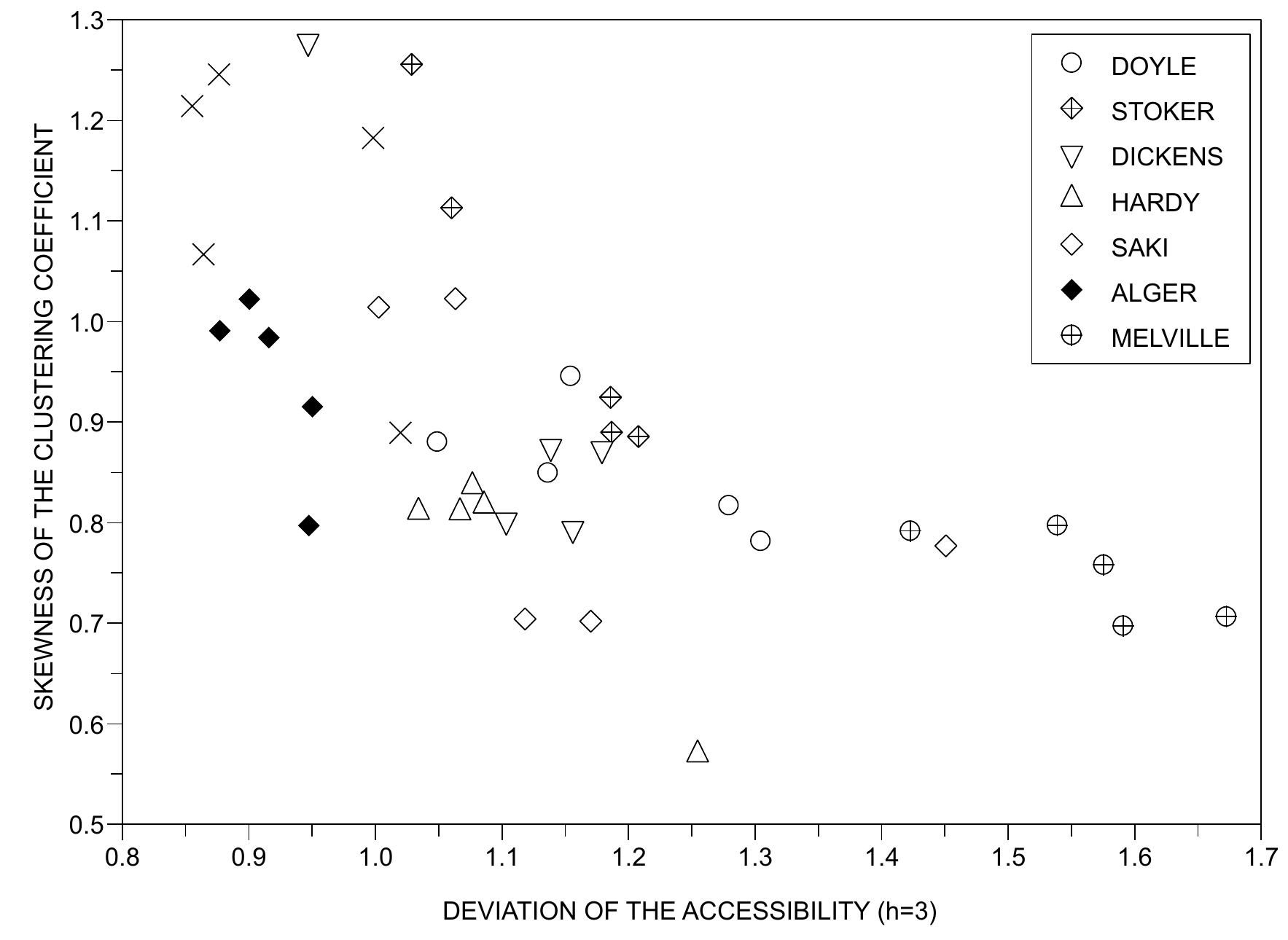}
  \caption{\label{fig.sep}Discriminability of authors obtained with two topological features of complex networks modelling texts. Note that, using only two features, it was possible to separate e.g. Alger from Melville. According to the information gain criterion, the most relevant network features for the authorship identication task were the standard deviation of the accessibility computed at the third level ($\Omega(\Delta \alpha^{(h=3)}) = 1.07$) and the \Red{standard deviation of the average neighboorhood degree ($\Omega(\Delta k^{(n)}) = 1.05$)}.  }
\end{figure}

The results in tables \ref{tab.lambda.aut}-\ref{tab.tie.out} confirms that the topology of networks modeling texts is able to improve the characterization of texts because such textual description grasps relevant patterns that are mostly disregarded by traditional statistical approaches. \Red{This can be confirmed by optimized results obtained for $\lambda^*>0$ and $\theta^*>0$}.
To better understand the factors \Red{behind} the discriminability power of the network model, the relevance of each topological for discriminating authors was computed with the information gain criterion. The best topological features \Red{were the standard deviation of the accessibility at the third level} ($\Omega(\Delta \alpha^{(h=3)}) = 1.07$), the standard deviation of the \Red{average neighboorhood degree ($\Omega(\Delta k^{(n)}) = 1.05$)} and the \Red{average clustering coefficient} ($\Omega(\langle C \rangle) = 0.60$). An example of discriminability provided by the topological strategy is shown in figure \ref{fig.sep}.

%
%
%

\subsection{Style identification} \label{style.id}

In this section, I investigate if the topological features of complex  networks are useful to complement the traditional textual characterization for the style identification task~\cite{itask}. The dataset used was the Brown corpus~\cite{bcorpus}, which comprises documents that are classified either as informative prose (e.g. press reportage, popular folklore and scientific manuscripts) or imaginative prose (e.g general fiction, romance and love stories). Differently from the authorship attribution task, the objective here is to cluster together documents written using the same style, regardless of their authorship.

\begin{figure}
  \centering
  \includegraphics[width=0.45\textwidth]{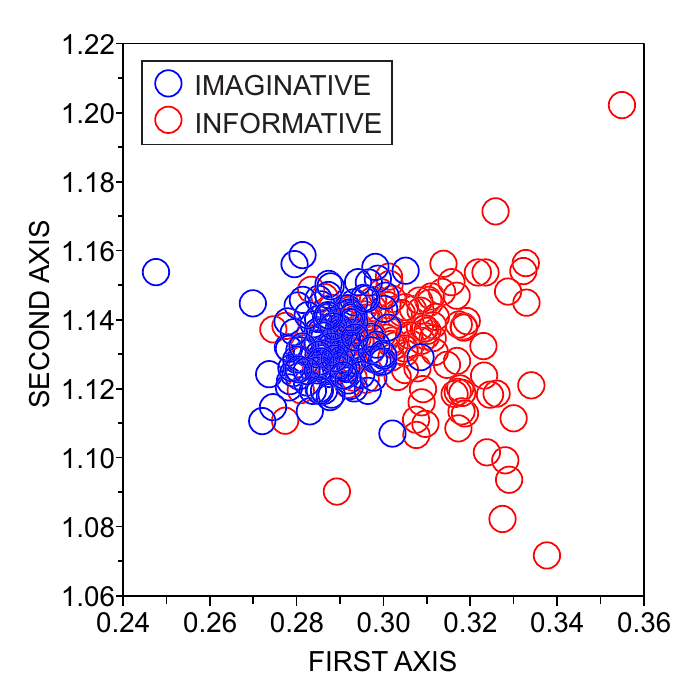}
  \caption{\label{fig.fig}Projection of the Brown dataset using topological features of networks modeling texts. The linear discriminant analysis~\cite{ldaref} was employed to generate the visualization. Note that the variability of the documents classified as imaginative prose is lower than the variability of style observed for informative documents.}
\end{figure}

The results obtained for the style identification in the Brown corpus using the hybrid technique are shown in tables \ref{abd.tab} and \ref{abd2.tab}. Considering the INT+CN combination, \Red{the highest gain in performance were 29.0\% and 25.8\%} for the kNN ($k=4$) and SVM methods. Minor improvements in accuracy were observed for other combinations, which might be explained by the high discriminability rates observed for the traditional techniques based on the frequency of stopwords ($\Gamma=95.6\%$) and character bigrams ($\Gamma = 94.05\%$). The results obtained with the tiebreaker technique were inferior to those obtained with the hybrid technique (results not shown). All in all, the results confirm that the inclusion of topological attributes might also improve the characterization of documents in the context of stylistic-based classification tasks. In this task, the most relevant topological features were the skewness and deviation of the clustering coefficient ($\Omega(\gamma(C)) = 0.172$ and $\Omega(\Delta C) = 0.157$). A visualization of the discriminability provided by topological features is shown in figure~\ref{fig.fig}. %

\begin{table}[h]
\centering
\caption{\label{abd.tab}Best relative accuracy rate $\Gamma_{max}$ obtained with the hybrid classifier based on the k-nearest neighbors method. The threshold employed $\lambda^*$ to obtain the best accuracy rate is also shown.}
\begin{tabular}{|l|cc|cc|cc|cc|cc|}
\hline
 & \multicolumn{2}{|c|}{$k=1$} & \multicolumn{2}{|c|}{$k=2$} & \multicolumn{2}{|c|}{$k=3$} &\multicolumn{2}{|c|}{$k=4$} & \multicolumn{2}{|c|}{$k=5$} \\
 \cline{2-11}
 & $\Gamma_{max}$ & $\lambda^*$ & $\Gamma_{max}$ & $\lambda^*$ & $\Gamma_{max}$ & $\lambda^*$ & $\Gamma_{max}$ & $\lambda^*$ & $\Gamma_{max}$ & $\lambda^*$ \\
\hline
{\bf Intermittence} & 1.127 & 0.50 & 1.232 & 0.55 & 1.05 & 0.48 & 1.290 & 0.76 & 1.000 & -- \\
{\bf Stopwords}     & 1.000 & -- & 1.010 & 0.01 & 1.013 & 0.25 & 1.017 & 0.34 & 1.021 & 0.38 \\
{\bf Characters}    & 1.000 & -- & 1.000 & -- & 1.000 & -- & 1.019 & 0.39 & 1.025 & 0.32 \\
\hline
\end{tabular}
{}
\end{table}

\begin{table}[h]
\centering
\caption{\label{abd2.tab}Best relative accuracy rate $\Gamma_{max}$ obtained with the tiebreaker algorithm applied to the Support Vector Machine (SVM), Random Forest (RFO) and Multi Layer Perceptron (MLP) methods.}
\begin{tabular}{|l|cc|cc|cc|}
\hline
 & \multicolumn{2}{|c|}{{\bf SVM}} & \multicolumn{2}{|c|}{\bf RFO} & \multicolumn{2}{|c|}{\bf MLP} \\
 \cline{2-7}
 & $\Gamma_{max}$ & $\theta^*$ & $\Gamma_{max}$ & $\theta^*$ & $\Gamma_{max}$ & $\theta^*$ \\
\hline
{\bf Intermittence} & 1.258 & 0.51 & 1.111 & 0.63 & 1.015  & 0.29 \\
{\bf Stopwords}     & 1.000 & -- & 1.000 & --   & 1.000  & -- \\
{\bf Characters}    & 1.000 & -- & 1.000 & --   & 1.000  & -- \\
\hline
\end{tabular}
{
}
\end{table}

\section{Conclusions} \label{aconclusao}

In this study, I have shown that the styles of texts can be captured by measuring the topological properties of the corresponding network representation. More important than just to note a dependency between the structure of networks and stylistic features of texts, this study devised techniques to combine traditional and topological features, which were found to be of paramount importance to enhance the quality of current classification strategies. Two traditional stylometry tasks were studied: the authorship attribution and genre identification problems. In both tasks, the addition of topological features provided an improvement in classification performance. The highest improvement occurred with the hybrid classifier, which uses a linear convex combination of features from distinct textual evidences. The different nature of the quantities used to characterize texts suggests a complementary role in capturing distinct aspects of written texts.
\Red{Because the adequate choice of parameters for the proposed technique is far from being a trivial task, I intend, as a future work, to devise a method that automatically assigns a value of $\lambda$ and $\Theta$ for each test instance.}
%

%
%

\section*{Acknowledgments}
I acknowledge financial support from S\~ao Paulo Research Foundation (FAPESP-Brazil)
(grant number 14/20830-0).


\section*{References}

\begin{thebibliography}{10}

\bibitem{zipf0}
Zipf GK (1949)
{Human Behavior and the Principle of Least Effort}.
Addison-Wesley

\bibitem{zipf}
Kali R (2003) The city as a giant component: a random graph approach to Zipf's law. Appl. Econ. Lett. 10 (4) 717--720.

\bibitem{zipf1}
Baixeries J, Elvevag B, Ferrer-i-Cancho R (2013)
The evolution of the exponent of Zipf's law in language ontogeny.
{PLoS ONE} {8} e53227.

\bibitem{heap1}
Bochkarev VV, Lerner EY, Shevlyakova AV (2014)
Deviations in the zipf and heaps laws in natural languages.
{J. Phys.: Conf. Ser.} {490} 012009.
%

\bibitem{heap2}
Font-Clos F, Boleda G, Corral A (2013)
A scaling law beyond Zipf's law and its relation to Heaps' law.
{New J. Phys.} {15} 093033.

\bibitem{heap3}
Lu L, Zhang Z, Zhou T (2013)
Deviation of Zipf's and Heaps' laws in human languages with limited dictionary sizes.
{Sci. Rep.} {3} 1082.

\bibitem{katz}
Katz SM (1996)
Distribution of content words and phrases in text and language modelling.
{Nat. Lang. Eng.} {2} 15--59.
%

\bibitem{authorship}
Amancio DR (2015)
Authorship recognition via fluctuation analysis of network topology and word intermittency.
{J. Stat. Mech.} P03005.
%

\bibitem{level}
Carpena P, Bernaola-Galvan P, Hackenberg M, Coronado AV, Oliver JL (2009)
Level statistics of words: Finding keywords in literary texts and symbolic sequences.
{Phys. Rev. E} {79} 3.

\bibitem{statistical}
Herrera JP, Pury PA (2008)
Statistical keyword detection in literary corpora.
{EPJ B} {63} 135--146.

\bibitem{imeas}
Allegrini P, Grigolini P, Palatella L (2004)
Intermittency and scale-free networks: a dynamical model for human language complexity.
{Chaos Soliton Fract.} {20} 95.

\bibitem{ref1}
Kulig A, Drozdz S, Kwapien J, Oswiecimka P (2015)
Modeling the average shortest-path length in growth of word-adjacency networks.
{Phys. Rev. E} {91} 032810.

\bibitem{wan}
Roxas RM, Tapang G (2010)
Prose and poetry classification and boundary detection using word adjacency network analysis.
{Int. J. Mod. Phys. C} { 21} 503--512.

\bibitem{masucci}
Masucci AP, Rodgers GJ (2006)
Network properties of written human language.
{Phys. Rev. E} {74} 026102.

\bibitem{hliu}
Cong J, Liu H (2014)
Approaching human language with complex networks.
{Phys. Life Rev.}  {11} 4.
%


\bibitem{origins}
Ferrer i Cancho R, Sole RV (2003)
Least effort and the origins of scaling in human language.
{Proc. Natl. Acad. Sci. USA} {100} 788--91.

\bibitem{acquisition}
Corominas-Murtra B, Balverde S, Sol\'e R (2009)
The ontogeny of scale-free syntax networks: phase transitions in early language acquisition.
{Advs. Complex Syst.} 12(3) 371--392.

\bibitem{cnammt}
Amancio DR, Antiqueira L, Pardo TAS, Oliveira Jr. ON, Nunes MGV (2008)
Complex networks analysis of manual and machine translations.
{Int. J. Mod. Phys. C} {19} 583.

\bibitem{extractive}
Amancio DR, Nunes MGV, Oliveira Jr. ON, Costa LF (2012)
Extractive summarization using complex networks and syntactic dependency.
{Physica A} {391} 1855--1864.

\bibitem{unveiling}
Amancio DR, Oliveira Jr. ON, Costa LF  (2012)
Unveiling the relationship between complex networks metrics and word senses.
{Europhys. Lett.} {98} 18002.


\bibitem{navigli}
Navigli R (2009)
Word sense disambiguation: a survey.
{ACM Comput. Surv.} { 41} 2.

\bibitem{complexity}
Amancio DR, Aluisio SM, Oliveira Jr. ON, Costa LF (2012)
Complex networks analysis of language complexity.
{Europhys. Lett.} {100} 58002.

\bibitem{forte}
Antiqueira L, Nunes MGV, Oliveira Jr ON, Costa LF (2007)
Strong correlations between text quality and complex networks features.
{Physica A} {373} 811--820.

\bibitem{mihalcea}
Mihalcea R, Radev D (2012)
{Graph-based natural language processing and information retrieval}.
Cambridge University Press, isbn 978-0-521-89613-9.

\bibitem{surveyaut}
Stamatatos E (2009)
A survey of modern authorship attribution methods.
{J. Am. Soc. Inf. Sci. Technol.} {60} 3 538--556.

\bibitem{liter}
Yu B (2006)
{An Evaluation of Text Classification Methods for Literary Study}.
Ph.D. Dissertation. University of Illinois at Urbana-Champaign, Champaign, IL, USA.

\bibitem{fuzzyknn}
Keller JM, Gray MR, Givens Jr. JA (1985)
A fuzzy k-nearest neighbor algorithm.
{IEEE Trans. Syst. Man, Cybern.} {15} 4 580--585.

\bibitem{baroncheli}
Baronchelli A, Ferrer-i-Cancho R, Pastor-Satorras R, Chater N, Christiansen MH (2013)
Networks in cognitive science.
{Trends Cogn. Sci.} {17} 348--60.

\bibitem{lantiq}
Antiqueira L, Oliveira Jr. ON, Costa LF, Nunes MGV (2009)
A complex network approach to text summarization,
{Inform. Sci.} {179} 584--599.

\bibitem{patterns}
Ferrer i Cancho R, Sol\'e R, Kohler R (2004)
Patterns in syntactic dependency networks.
{Phys. Rev. E} {69} 51915.

\bibitem{arenas}
Capitan JA, Borge-Holthoefer J, Gomez S, Martinez-Romo J, Araujo L, Cuesta JA, Arenas A (2012)
Local-based semantic navigation on a networked representation of information.
{PLoS ONE} {7} 8 e43694.

\bibitem{adv}
Gravino P, Servedio VDP, Barrat A, Loreto V (2012)
Complex structures and semantics in free word association.
{Adv. Complex Syst.} {15} 1250054--1.
%

\bibitem{highorder}
Silva TC, Amancio DR (2012)
Word sense disambiguation via high order of learning in complex networks.
{Europhys. Lett.} {98} 58001.

\bibitem{veronis}
Veronis J (2004)
Hyperlex: Lexical cartography for information retrieval.
{Comput. Speech Lang.} {18} 3 223--252.

\bibitem{windows}
Widdows D, Dorow B (2002)
{A graph model for unsupervised lexical acquisition}.
{Proceedings of the 19th International Conference on Computational Linguistics}  1--7.

\bibitem{disen}
\Red{
Martinez-Romo J, Araujo L, Borge-Holthoefer J, Arenas A, Capitan JA, Cuesta JA (2011) Disentangling categorical relationships through a graph of co-occurrences. Phys. Rev. E 84 046108.
}

\bibitem{identification}
Amancio DR, Oliveira Jr. ON, Costa LF (2012)
Identification of literary movements using complex networks to represent texts.
{New J. Phys.} {14} 043029.

\bibitem{machine}
Amancio DR, Nunes MGV, Oliveira Jr. ON, Pardo TAS, Antiqueira L, Costa LF (2011)
Using metrics from complex networks to evaluate machine translation.
{Physica A} {390} 131--142.

\bibitem{voynich}
\Red{
Amancio DR, Altmann EG, Rybski D, Oliveira Jr. ON, Costa LdF (2013)
Probing the statistical properties of unknown texts: application to the Voynich manuscript. PLoS ONE 8 e67310.
}

\bibitem{verref}
Berger AL, Della Pietra VJ, Della Pietra SA (1996)
A maximum entropy approach to natural language processing.
{Comput. Linguist.} {22} 1 39--71.

\bibitem{surveymeas}
Costa LF, Rodrigues FA, Travieso G, Villas Boas PRV (2007)
Characterization of complex networks: a survey of measurements.
{Adv. Phys.} {56} 167--242.


\bibitem{58}
Antiqueira L, Pardo TAS, Nunes MGV, Oliveira Jr. ON (2007)
Some issues on complex networks for author characterization.
{\it Inteligencia Artificial} {11} 51--58.








\bibitem{travencolo}
Travencolo B, Costa LF (2008)
Accessibility in complex networks.
{Phys. Lett. A} {373} 89--95.

\bibitem{concentrico}
Costa LF, Tognetti MAR, Silva FN (2008)
Concentric characterization and classification of complex network nodes: application to an institutional collaboration network.
{Physica A} {387} 6201--6214.

\bibitem{probing}
Amancio DR (2015)
Probing the topological properties of complex networks modeling short written texts.
{PLoS ONE} {10} e0118394.

\bibitem{commrev}
Fortunato S (2010)
Community detection in graphs.
{Phys. Rep.} {486} 75--174.
%

\bibitem{comparing}
Amancio DR, Altmann EG, Oliveira Jr. ON, Costa LF (2011)
Comparing intermittency and network measurements of words and their dependence on authorship.
{New J. Phys.} {13} 123024.

\bibitem{mixing}
Newman MEJ (2003)
Mixing patterns in networks.
{Phys. Rev. E} {67} 2 026126.


\bibitem{manning}
Manning CD, Schutze H (1999)
{Foundations of Statistical Natural Language Processing}. MIT Press, Cambridge, MA, USA.

\bibitem{mosteller}
Mosteller F, Wallace D (1964).
Inference and Disputed Authorship: The Federalist.
Reading, MA: Addison-Wesley.

\bibitem{28}
Havlin S (1995)
The distance between zipf plots.
{Physica A} {216} 148--50.

\bibitem{29}
Vilensky B (1996)
Can analysis of word frequency distinguish between writings of different authors?
{Physica A} {231} 705--11.

\bibitem{30}
Yang ACC, Peng CK, Yien WK, Goldberger AL (2003)
Information categorization approach to literary authorship disputes.
{Physica A} {329} 473--83.



\bibitem{52}
Jankowska M, Keselij V, Milios E (2012)
Relative n-gram signatures: document visualization at the level of character n-grams.
{IEEE Conference on Visual Analytics Science and Technology}, Seattle, WA, USA.

\bibitem{rtimes}
Goh KI, Barab\'asi AL (2008)
Burstiness and memory in complex systems.
{Europhys. Lett.} {81} 48002.
%


\bibitem{wekabook}
Witten IH, Frank E (2005)
{Data Mining: Practical Machine Learning Tools and Techniques}.
Morgan Kaufmann, San Francisco.

\bibitem{infogain}
Mitchell TM (1997)
{Machine Learning}.
The Mc-Graw-Hill Companies, Inc. ISBN 0070428077.



\bibitem{fsvm}
\Red{
Lin C-F abd Wang S-D (2002)
Fuzzy support vector machines.
IEEE Trans. Neural Netw. Learn. Syst. 13(2) 464--471.
}

\bibitem{ftree}
\Red{
Olaru C and Wehenkel L (2003)
A complete fuzzy decision tree technique.
Fuzzy Set. Syst. 138(2) 221--254.
}

\bibitem{fforest}
\Red{
Bonissone P, Cadenas JM, Garrido MC and Díaz-Valladares RA (2010)
A fuzzy random forest.
Int. J. Approx. Reason. 51(7) 729--747.
}

\bibitem{fmlp}
\Red{
Pal SK and Mitra S (1992)
Multilayer perceptron, fuzzy sets, and classification.
IEEE Trans. Neural Netw. Learn. Syst. 3(5) 683--697.
}

\bibitem{grid1}
\Red{
Bergstra J and Bengio Y (2012)
Random search for hyper-parameter optimization.
J. Machine Learning Research 13: 281--305.
}

\bibitem{grid2}
\Red{
Bergstra J, Bardenet R, Bengio Y and Kegl B (2011)
Algorithms for hyper-parameter optimization.
Adv. Neural Inform. Process. Syst.
}

\bibitem{systematic}
Amancio DR, Comin CH, Casanova D, Travieso G, Bruno OM, Rodrigues FA and Costa LF (2014)
A systematic comparison of supervised classifiers.
{PLoS ONE} {9} e94137.



\bibitem{t5}
Oakes M (2004)
Ant colony optimisation for stylometry: the federalist papers.
{Proc. 5th Int. Conf. on Recent Advances in Soft Computing}.


\bibitem{itask}
Patton JM, Can F (2004)
A stylometric analysis of Yasar Kemal's Ince Memed tetralogy.
Computers and the Humanities 38 (4) 457--467.

\bibitem{bcorpus}
Available from http://clu.uni.no/icame/manuals/BROWN/INDEX.HTM

\bibitem{ldaref}
McLachlan GJ (2004)
{Discriminant Analysis and Statistical Pattern Recognition}.
Wiley Interscience. ISBN 0-471-69115-1.

\end{thebibliography}
\end{document}